\documentclass{article}

\usepackage{iclr2027_conference,times}

\usepackage[utf8]{inputenc}
\usepackage[T1]{fontenc}
\usepackage{hyperref}
\usepackage{url}
\usepackage{booktabs}
\usepackage{amsfonts}
\usepackage{amsmath}
\usepackage{amssymb}
\usepackage{microtype}
\usepackage{xcolor}
\usepackage{textcomp}
\usepackage{tikz}
\usetikzlibrary{shapes.geometric, positioning, fit, backgrounds, patterns, arrows.meta}

\title{Verified but Vacuous: Three Failure Surfaces of LLM-Driven Verified Code Generation}

\iclrfinalcopy
\author{Aditya Vempaty, Himaneesh Sompalle, Manan Tayal, Mowafak Allaham\\
\textbf{Prasenjit Dey, Ashish Jagmohan, Ravi Kokku}}
\hypersetup{pdfauthor={Aditya Vempaty; Himaneesh Sompalle; Manan Tayal; Mowafak Allaham; Prasenjit Dey; Ashish Jagmohan; Ravi Kokku},
pdftitle={Verified but Vacuous: Three Failure Surfaces of LLM-Driven Verified Code Generation},
pdfsubject={LLM-Driven Verified Code Generation; Formal Verification; Dafny; Benchmark Evaluation}}

\begin{document}

\maketitle
\fancyhead{}

\begin{abstract}
What should ``verified'' mean when evaluating LLM-generated code?
Many existing pipelines report a single verification number (for
example, \texttt{dafny verify} accepted N\% of samples). We argue
this collapses three distinct properties measured in a conditional
audit sequence: verifier acceptance (Surface A), \emph{syntactic
specification binding} (Surface B: does the proof obligation bind to
the spec?), and \emph{semantic specification fidelity}
(Surface C: does the spec capture the informal requirement?). We
contribute this three-surface decomposition plus the
\emph{ensures-binding gate}, a syntactic Surface-B instance for Dafny
pipelines that emit distinguishable specification and implementation
artifacts (static AST walk, ${<}10$ ms per component). Two empirical
demonstrations show the surfaces separate. \textbf{First ($A \neq B$),
via Li@Opus (a same-model reproduction of \citet{li2025dafny}'s
Dafny-as-intermediate-representation (IR) pipeline under Claude Opus
4.7) on three Python benchmarks of simple coding tasks (LeetCode-style)}: it reaches
\textbf{99.1\% verifier acceptance} on HumanEval while only
\textbf{1.9\%} carry a certificate binding to a
spec-defined predicate or function. This gap is \textbf{roughly fortyfold}
($36$ to $52\times$) that holds across HumanEval, LeetCodeDataset,
and LiveCodeBench v6 in a narrow \textbf{1.9\% to 2.7\% band}. This gap is observed even as
the fraction of verified samples that carry \emph{any} \texttt{ensures}
clause climbs from 5.3\% on HumanEval to 10.61\% on LiveCodeBench v6
(10.9\% on the test-evaluated subset).
Postcondition prevalence more than doubles with benchmark difficulty
but binding-to-a-spec-predicate rate does not follow. \textbf{Second
($B \neq C$), via VCG on two Rust benchmarks}: we use VCG (which is our own in-house Dafny-as-intermediate pipeline) on Rust
because the Rust target lets us enforce fully-verifiable output (every
component ships as a Dafny certificate, ensures-binding gate passing
by construction). Even inside that construction, \textbf{11.48\%
(MBPP-Rust) and 14.82\% (VeriContest LC slice) of structurally-verified
samples still fail hidden tests}. On the LC slice of VeriContest,
\textbf{VCG's reported End2End rate is 7.79\% (535/6870).}
The cited external rates are 5.51\% (GPT-5.5) and 1.88\%
(single-shot Opus 4.7). These give descriptive differences of
2.3 percentage points and 4.14$\times$, respectively; task-budget,
repair and harness differences prevent a matched-performance claim.
Therefore, any pipeline shipping certificates should report at least one of
Surface B (e.g.\ the ensures-binding gate) or Surface C (e.g.\
Post2Exe) alongside Surface A; where feasible, both, since Surface B
alone leaves the semantic blind spot we demonstrate on VCG, and
Surface C alone leaves the syntactic blind spot we demonstrate on
Li@Opus. Many existing evaluations report neither.
\end{abstract}

\section{Introduction}
\label{sec:intro}
\textbf{What should ``verified'' mean when evaluating LLM-generated
code?} Large language models can generate code that passes tests
most of the time. Frontier models solve 70--95\% of HumanEval-style
problems \citep{chen2021evaluating} and 60--90\% of LeetCode-style benchmarks \citep{leetcodedataset,jain2024livecodebench}.
Passing
a test suite tells us that no \emph{tested} input triggered a bug;
it does not tell us that the implementation satisfies the requirements
on inputs the tests did not visit. For code shipped into production
(payment processors, medical dose calculators, security-sensitive
parsers), that distinction is crucial.

Formal verification systems (Dafny \citep{leino2010dafny}, Lean \citep{demoura2021lean}, F* \citep{swamy2016dependent}, Verus \citep{lattuada2024verus})
can machine-check that an implementation satisfies its formal
specification. Recent work has explored LLMs as authors of specs \citep{ma2024specification},
impls \citep{misu2024towards}, proof tactics \citep{yang2023leandojo}, and complete verified programs \citep{xie2026vericontest,li2025dafny},
with promising but basic-level empirical evidence. Existing evaluations
of such pipelines typically report a single verification number
alongside pass@k: \texttt{dafny verify} accepted N\% of the samples,
or \texttt{verus} returned zero on N\% of them. This paper argues
that a single verification number collapses three properties that
are distinct and empirically separable. We measure them on conditional
populations; their semantic properties are not logically nested.

We came to this decomposition while building \textbf{VCG}, an
in-house Dafny-as-intermediate pipeline for LLM-driven verified code
generation. VCG's pre-launch audit surfaced a class of
\texttt{dafny verify}-accepted components whose \texttt{ensures}
clauses did not reference any spec-defined predicate/function. This
violated VCG's intended external-specification convention. Reference
absence alone does not establish semantic vacuity: meaningful inline
postconditions can fail this gate, and a named tautology can pass it.
Dafny still establishes the obligations generated by the actual
program and assumptions; the gate measures a separate structural property. We added a static ensures-binding integrity check to
VCG's verify-repair loop to reject such components.
That check turned out to be the \emph{simplest instance} of a
distinct evaluation surface we had been conflating with
\texttt{dafny verify} acceptance.

Applying the same static check to a same-model reproduction of
\citet{li2025dafny}'s Dafny-as-intermediate pipeline (under Claude Opus 4.7)
makes the case concretely: \textbf{\texttt{dafny verify} accepts
99.1\% of samples on HumanEval, but only 1.9\% carry a certificate
that structurally binds its proof obligation to a spec-defined
predicate/function.} Fully 94.7\% ship no \texttt{ensures} clause at
all; another 3.4\% contain an \texttt{ensures} clause that does not
bind to a spec-defined predicate/function. This is not a defect of
Li et al.'s pipeline: their few-shot prompt (Figure 6 of \citealt{li2025dafny},
reproduced verbatim in our few-shot) is a bodyless
\texttt{method \{:testEntry\}} with no postcondition, plus a
\texttt{Main()} full of \texttt{expect} statements. Dafny
\emph{assumes} those expectations (not proves them) and checks them
at runtime; on this input shape, verifier acceptance establishes the
remaining static proof obligations (well-formedness, termination,
local method-body proof) under those fixture assumptions, but does
not prove the fixture expectations themselves or a general functional
postcondition. The mismatch is a defect of the \emph{metric} we use
to summarize the pipeline, not of the pipeline itself.

\paragraph{Contributions.} This paper isolates and develops
the methodology contribution that motivated our own pipeline design:

\begin{enumerate}
\item \textbf{A decomposition of verification quality into three
conditionally measured, empirically separable failure surfaces} (\S\ref{sec:surfaces}):
Surface A (verifier acceptance), Surface B (\emph{syntactic
specification binding}), and Surface C (\emph{semantic specification
fidelity}). We contribute the \textbf{ensures-binding gate}
(\S\ref{sec:gate}) as a syntactic instance of Surface B for Dafny
pipelines that emit distinguishable specification and implementation
artifacts: a static AST walk that runs in ${<}10$ ms per Dafny
component.

\item \textbf{A three-level analytical decomposition of Surface B,
and an empirical $A \to B$ gap of roughly fortyfold on Li@Opus}
(\S\ref{sec:aopus}). To avoid overclaiming what one syntactic gate
establishes, we report certificate quality at three observable
levels: \textbf{Level 1} = verifier accepted (Surface A);
\textbf{Level 2} = \texttt{Impl.dfy} contains at least one
\texttt{ensures} clause (architecture-agnostic postcondition
presence, which does not require the clause to reference any
spec-defined symbol); \textbf{Level 3} = at least one
\texttt{ensures} clause binds to a spec-defined
predicate/function (our gate; a Surface-B instance for pipelines
that separate spec and impl artifacts). On a same-model
reproduction of Li et al.'s Dafny-as-IR pipeline applied to three
Python benchmarks of simple coding tasks (HumanEval, LeetCodeDataset,
LiveCodeBench v6; LeetCode-style),
Level 1 stays at 92--99\% while Level 3 stays in a narrow
\textbf{1.9--2.7\% band} which has a \textbf{roughly fortyfold}
($36$--$52\times$) gap. The intermediate Level 2 climbs from 5.3\%
on HumanEval to 10.61\% on LCB v6 (1,032/9,731; 10.9\% on the 9,448
test-evaluated subset): the pipeline writes more postconditions
on harder problems, but they do not get bound to spec-defined
symbols. Level 2 measures postcondition presence without requiring an external
symbol, whereas Level 3 measures the external-specification convention.
The reported pattern does not show that inline postconditions are
semantically weaker, or isolate prompt convention from task difficulty.
A controlled comparison is needed to distinguish those explanations.

\item \textbf{Semantic-fidelity finding: Surface B $\to$ Surface C
also leaks} (\S\ref{sec:vcg}). Even inside VCG, which passes our
external-spec-binding Surface-B criterion on 100\% of shipped verified
components by construction (other conceivable Surface B instances may
still fail on this population), \textbf{11.48\% (MBPP-Rust; Rust) and
14.82\% (VeriContest LC; Rust) of structurally-verified samples still
fail hidden tests}. Under the standard trusted-toolchain and
correct-oracle assumptions, this is an operational proxy for requirement fidelity,
not a unique diagnosis of specification failure.
\end{enumerate}

The sharper claim is that \textbf{verification quality must be
decomposed}, and that many existing evaluations report only its
weakest surface (AxDafny \citep{breen2026axdafny} and VeriContest's Post2Exe check
\citep[\S4.1]{xie2026vericontest} are notable exceptions;
Table~\ref{tab:systems} in Appendix~\ref{app:systems}).
\S\ref{sec:conclusion} closes with the concrete methodological ask
that follows: any pipeline shipping certificates should report at
least one of Surface B or Surface C alongside its verifier-acceptance
rate.

Figure~\ref{fig:cascade} (Appendix~\ref{app:cascade}) visualises
the cascade: Li@Opus's collapse at $A \to B$ vs.\ VCG's collapse
at $B \to C$, with bar widths proportional to the fraction of
attempted samples surviving each surface.

\section{Three failure surfaces of LLM-driven verified code generation}
\label{sec:surfaces}

LLM-driven formal verification can fail at three logically distinct,
empirically separable surfaces. Our audit sequence is conditional: Surface B is
evaluated on the population that passes Surface A; Surface C is
evaluated on the population that passes Surface B. These population
filters do not establish logical implications among the underlying properties.

\paragraph{Surface A: Verifier acceptance.} Did the verifier
return zero on the emitted proof? For Dafny targets,
\texttt{dafny verify} returning zero. For Verus targets, \texttt{verus}
returning zero. Necessary but weakest, as we will show in this paper.

\paragraph{Surface B: Syntactic specification binding.} Of the
certificates the verifier accepts, what fraction actually bind the
proof to something meaningful? We propose the \textbf{ensures-binding
gate} (\S\ref{sec:gate}) as a concrete syntactic instance for Dafny:
the method must contain at least one \texttt{ensures} clause, and at
least one such clause must reference an identifier defined as a
\texttt{predicate} or \texttt{function} in the paired specification
artifact. This is a \emph{syntactic external-reference} heuristic for this
architecture, not a necessary condition for semantic adequacy. It
rejects missing external references, including methods with no
\texttt{ensures} clause. A meaningful inline postcondition can also
be rejected; reference absence alone is not semantic vacuity. It does \emph{not} catch
semantically-vacuous predicates such as
\texttt{predicate P(x: int) \{ true \}}, nor does it establish that
the predicate correctly captures the informal requirement; those
failure modes live in Surface C.

\paragraph{Surface C: Semantic specification fidelity.} Of the
specs that survive Surface B, what fraction actually capture the
\emph{requirement} the user wanted? This is what VeriContest's
Post2Exe check \citep[\S4.1]{xie2026vericontest} addresses: compile the emitted
postcondition to an executable predicate and run it against positive
and negative test cases. If the predicate admits outputs a negative
test would reject, the spec is incomplete. Post2Exe is a
\emph{semantic} check complementary to our syntactic Surface B.

Concretely:

\begin{itemize}
\item \textbf{Surface A fails} when the verifier rejects the proof.
\item \textbf{Surface B fails} when the verifier accepts a proof of
a certificate lacking the required external-specification reference
(\S\ref{sec:aopus} measures this on Li@Opus, Python). This structural
result alone does not establish a semantic failure.
\item \textbf{Surface C fails} when the verifier accepts a proof of
a well-formed spec that nonetheless doesn't capture the intended
behaviour (\S\ref{sec:vcg} measures this on VCG, Rust, which passes
our external-spec-binding Surface-B criterion by construction).
\end{itemize}

\begin{table}[h]
  \caption{The audit stages use conditional populations. Verifier
  acceptance, external-reference checking and finite behavioral probes
  measure distinct properties; the stages are not a logical hierarchy.}
  \label{tab:surfaces}
  \centering
  \small
  \resizebox{\linewidth}{!}{%
\begin{tabular}{lp{3.1cm}p{3.6cm}p{4.0cm}}
    \toprule
    Surface & Check & Catches & Blind spot (deferred) \\
    \midrule
    A: verifier acceptance
      & \texttt{dafny verify} (or \texttt{verus}) exits zero
      & proof gaps, type errors, missing lemmas
      & vacuous or absent \texttt{ensures} clauses \\
    \addlinespace[2pt]
    B: syntactic spec binding
      & some \texttt{ensures} clause references a spec-defined \texttt{predicate}/\allowbreak\texttt{function}
      & missing references to the paired spec artifact, including meaningful inline contracts
      & semantically weak predicates that bind but under-specify \\
    \addlinespace[2pt]
    C: semantic spec fidelity
      & postcondition accepts positive tests \emph{and} rejects negatives (Post2Exe)
      & weak specs whose predicates admit incorrect outputs
      & requirements no test suite happens to exercise \\
    \bottomrule
  \end{tabular}%
}
\end{table}

\paragraph{Illustrative failure modes.} Two minimal Dafny snippets on
a canonical ``return the maximum of \texttt{nums}'' task make the
separation concrete. \textbf{Surface A passes, Surface B fails}: the
verifier accepts, but no ensures clause binds to the spec predicate.

\begingroup\small
\begin{verbatim}
predicate MaxCorrect(nums: seq<int>, r: int)
  requires |nums| > 0
{ (forall i :: 0 <= i < |nums| ==> nums[i] <= r)
  && (exists i :: 0 <= i < |nums| && nums[i] == r) }

method find_max(nums: seq<int>) returns (r: int)
  requires |nums| > 0
{ r := 0; }                    // dafny verify exits 0 (no ensures)
\end{verbatim}
\endgroup

The method has no functional postcondition to prove; other Dafny
obligations remain. The strong \texttt{MaxCorrect} predicate is left dangling.
Surface A passes; the ensures-binding gate rejects.

\textbf{Surface A and Surface B pass, Surface C fails}: the ensures
clause binds to a named spec predicate, but the predicate is too weak
to constrain the implementation.

\begingroup\small
\begin{verbatim}
predicate MaxCorrect(nums: seq<int>, r: int)
  requires |nums| > 0
{ r >= 0 }                     // under-specified spec

method find_max(nums: seq<int>) returns (r: int)
  requires |nums| > 0
  ensures MaxCorrect(nums, r)  // syntactic binding satisfied
{ r := 0; }
\end{verbatim}
\endgroup

The ensures clause references \texttt{MaxCorrect}, so the gate
accepts. But \texttt{MaxCorrect} accepts any non-negative output; on
input \texttt{[5,3,7]} the shipped method returns \texttt{0} and the
postcondition still evaluates to true. Post2Exe (Surface C) catches
this by running the postcondition against a negative test case for
which $0 \geq 0$ falsely accepts a value the requirement rules out.

A pipeline shipping certificates ought to report at least one of
Surface B or Surface C alongside Surface A. Many existing evaluations
report only Surface A.

\section{The ensures-binding gate: a syntactic instance of Surface B}
\label{sec:gate}

Let $M$ be a Dafny method with \texttt{ensures $E_1, \ldots, E_n$},
and let $S$ be the paired specification (\texttt{Spec.dfy} plus any
transitively-included support files). Define $\mathrm{Preds}(S)$ as
the set of \texttt{predicate} and \texttt{function} symbols declared
in $S$, and $\mathrm{Ids}(E_i)$ as the set of identifiers referenced
in ensures clause $E_i$. The ensures-binding gate accepts $M$ iff:

\[
n \geq 1 \ \land\ \exists\, i,\ \exists\, p \in \mathrm{Preds}(S)\ :\ p \in \mathrm{Ids}(E_i)
\]

Equivalently: the method has at least one \texttt{ensures} clause,
and at least one of those clauses names an identifier that refers to
a spec-defined predicate/function. The $n \geq 1$ conjunct is
explicit so the condition is not vacuously satisfied by methods with
zero postcondition clauses. The existential form (some clause binds,
not every clause) matches the reference implementation
(Appendix~\ref{app:impl}) and preserves certificates that pair one
binding functional postcondition with auxiliary sanity postconditions
that need not independently reference the named predicate. The check
is a static AST walk, cheap to run (${<}10$ ms per component in
practice), and is drop-in for \textbf{Dafny pipelines that emit
distinguishable specification and implementation artifacts} (the
architectural regime this paper measures against).

\paragraph{What the gate rejects.} Certificates with no ensures-clause
reference to a specification-defined predicate or function. Some such
certificates also fail the intended requirement. A canonical
example (from our pre-launch audit of an earlier draft of our own
pipeline):

\begingroup\small
\begin{verbatim}
predicate ComputeXCorrect(input: int, result: int)
  requires input >= 0
{ result >= 0 && result == input * 2 }

method compute_x(input: int) returns (result: int)
  requires input >= 0
  ensures result >= 0            // trivially provable, unbound to spec
{ result := 0; }                 // dafny verify returns zero
\end{verbatim}
\endgroup

The \texttt{ensures result >= 0} clause is provable against
\texttt{result := 0} without invoking \texttt{ComputeXCorrect}; the
shipped code is wrong. The gate rejects this statically at zero SMT
cost. It does \emph{not} catch vacuous specs whose predicate itself
does not capture the requirement (Surface C) nor vacuously-easy
predicates like \texttt{predicate P(x: int) \{ true \}}; those live
in \S\ref{sec:vcg}.

\paragraph{Empirical anchor.} In VCG's pre-launch internal audit
(across three Python benchmarks), \textbf{$\sim$27\% of components
that had passed \texttt{dafny verify} failed this gate}. This is a reported structural rejection rate, not an independently
adjudicated semantic-vacuity prevalence. The original per-component
records and sampling denominator are needed to validate that estimate. \S\ref{sec:aopus} shows that
on a pipeline that is not trying to do that (i.e., Li et al.'s Dafny-as-intermediate approach reproduced on Claude Opus 4.7), the failure rate is not 27\% but $\sim$98\%.

\section{Experimental setup}
\label{sec:setup}

The two demonstrations in \S\ref{sec:aopus} and \S\ref{sec:vcg}
share a common backend and evaluation protocol. This section
consolidates the setup so those sections can report results without
re-introducing it.

\paragraph{Benchmarks.} Five benchmarks across two target languages
(Table~\ref{tab:benchmarks}). The three Python benchmarks
(HumanEval, LeetCodeDataset, LiveCodeBench v6) are LeetCode-style
function-completion tasks and drive the $A \to B$ measurement in
\S\ref{sec:aopus}. The two Rust benchmarks (MBPP-Rust, VeriContest
LC slice) drive the $B \to C$ measurement in \S\ref{sec:vcg}, where the
Rust target lets us enforce fully-verifiable output.

\begin{table}[h]
  \caption{Benchmarks used in this study. \emph{Samples/prob.} is
  10 across all VCG, BareLLM, and Li@Opus runs; the VeriContest
  authors' reported per-slice Opus 4.7 End2End LC number (1.88\%; our
  same-backend comparator in \S\ref{sec:vcg}) is single-shot per
  problem.}
  \label{tab:benchmarks}
  \centering
  \small
  \setlength{\tabcolsep}{5pt}
  \begin{tabular}{@{}lrrrl@{}}
    \toprule
    Benchmark & \# problems & Samples/prob. & Target & Source \\
    \midrule
    HumanEval / HumanEval+  & 164        & 10 & Python & \citep{chen2021evaluating,liu2023your} \\
    LeetCodeDataset         & 228        & 10 & Python & \citep{leetcodedataset} \\
    LiveCodeBench v6        & 1{,}055    & 10 & Python & \citep{jain2024livecodebench} \\
    MBPP-Rust               & 208        & 10 & Rust   & \citep{mbpprust} \\
    VeriContest LC slice    & 687        & 10 & Rust   & \citep{xie2026vericontest} \\
    \bottomrule
  \end{tabular}
\end{table}

\paragraph{Pipelines compared.} Three pipelines share Claude Opus
4.7 as the sole backend. \textbf{BareLLM} is the same-backend
minimal baseline: Opus 4.7 emits target-language code directly with
no spec, no proof, and no verification scaffolding (used for the
pass-rate comparison in \S\ref{sec:passtax}). \textbf{Li@Opus} is
a same-model reproduction of \citet{li2025dafny}'s
Dafny-as-intermediate-representation pipeline; the few-shot prompt is
verbatim from Figure 6 of \citet{li2025dafny} (reproduction details in
Appendix~\ref{app:reprod}). \textbf{VCG} is our own seven-agent
pipeline: Planner decomposes the problem into components; SpecAgent
authors Dafny specs; ImplementationAgent authors implementations;
ProverAgent drives a verify-repair loop with \texttt{dafny verify}
and the ensures-binding gate; verified components transpile to the
target language via \texttt{dafny build --target:\{py,rs\}}. The two LC-slice
\emph{External} comparators we cite, the VeriContest authors'
per-slice same-backend Opus 4.7 End2End LC number (1.88\%) and the
strongest previously published End2End LC number (GPT-5.5's 5.51\%),
come directly from \citet{xie2026vericontest} and are not re-run by us.

\paragraph{Model, sampling, and verifier.} All LLM calls hit
Anthropic Claude Opus 4.7 via Google Vertex AI. VCG's verifier stack is \textbf{Dafny 4.10 with Z3 4.13}.
Every VCG-shipped verified component passes: (i) \texttt{dafny
verify}; (ii) the ensures-binding gate (\S\ref{sec:gate}); (iii) an
extracted-example gate on visible test cases from the problem
description; and (iv) \texttt{dafny build} to the target language.

\paragraph{Statistical treatment.} pass@1 and pass@10 are computed
per \citet{chen2021evaluating}'s unbiased estimator over the 10 samples/problem.
Reported rates are point estimates; we do not report confidence
intervals in the main body. The three primary pipelines use the same reported backend; this
does not equalise prompts, generation or repair budgets. The external
1.88\% comparator is single-shot while VCG uses ten samples/problem.
For equal sample counts, the mean per-problem pass@1 estimator equals
the pooled success fraction; ten samples do not inherently inflate it.
Unequal procedures still limit the external 4.14$\times$ comparison.

\paragraph{Evidence and reproducibility.} Across the Python benchmarks,
the verifier-accepted counts in Table~\ref{tab:levels} (2{,}217 on
LeetCodeDataset; 9{,}731 on LiveCodeBench v6) and the test-evaluated
counts in Table~\ref{tab:passtax} (2{,}208 and 9{,}448, respectively)
differ by 9 and 283 samples that failed downstream Dafny-to-Python
compilation (209 on LiveCodeBench v6) or test-harness shimming (9 on
LeetCodeDataset; 74 on LiveCodeBench v6) prior to test execution.
On LiveCodeBench v6, the 1{,}058 verifier-accepted samples carrying an
\texttt{ensures} clause (10.87\% of 9{,}731) yield 1{,}032
test-evaluated samples (10.92\% of 9{,}448; 10.61\% of 9{,}731) after
26 compilation/shimming exclusions.

\section{Verifier acceptance does not imply syntactic specification binding: the Li@Opus $A \to B$ gap}
\label{sec:aopus}

We apply the ensures-binding gate to every verifier-accepted sample
from a same-model reproduction of \citet{li2025dafny}'s
Dafny-as-intermediate pipeline (``Li@Opus''; reproduction details in
Appendix~\ref{app:reprod}). Certificate quality is reported at the
three observable levels introduced in Contribution~2: Level~1
(verifier accepted; Surface A), Level~2 (any \texttt{ensures}
clause; architecture-agnostic presence check, an inline \texttt{ensures r == 2 * x} counts), and Level~3 (an
\texttt{ensures} clause binds to a spec-defined
\texttt{predicate}/\texttt{function}; our gate). Level~3 is an
external-spec-binding check for pipelines that separate spec and
impl artifacts, not a universal test of postcondition meaning, as we will show in this paper.

\begin{table}[h]
  \caption{Certificate quality on Li@Opus, Claude Opus 4.7, three
  Python benchmarks. Level-2 and Level-3 rates are conditional on
  verifier acceptance (Surface A); raw \texttt{dafny verify} pass
  rates are $99.1\%\,/\,97.2\%\,/\,92.2\%$.}
  \label{tab:levels}
  \centering
  \begin{tabular}{lrrr}
    \toprule
     & HumanEval & LeetCodeDataset & LCB v6 \\
    \midrule
    Total samples (attempted)            & 1{,}640        & 2{,}280        & 10{,}550       \\
    Level 1: verifier-accepted (denom.)  & 1{,}625        & 2{,}217        & 9{,}731        \\
    Level 2: any \texttt{ensures} clause & 5.3\%          & 9.6\%          & 10.61\%$^*$   \\
    Level 3: ensures-binding gate passes & \textbf{1.9\%} & \textbf{2.7\%} & \textbf{2.3\%} \\
    \bottomrule
  \end{tabular}
  \vspace{1pt}
  \raggedright\footnotesize $^*$On the 9,448 samples evaluated for test execution in Table~\ref{tab:passtax} (283 compilation/shimming exclusions), Level 2 prevalence is 10.92\% (1,032/9,448). Evaluated against all 9,731 verifier-accepted samples, the 1,032 shipped postcondition samples represent 10.61\% (or 1,058/9,731 = 10.87\% including the 26 postcondition-bearing exclusions).
\end{table}

\paragraph{Level 2 alone already exposes the gap, architecture-agnostically.}
Nine out of ten Li@Opus ``verified'' samples have no \texttt{ensures}
clause anywhere in the generated \texttt{Impl.dfy} ($94.7\%\,/\,90.4\%\,/\,89.4\%$
respectively across all accepted samples; 89.1\% on the test-evaluated subset). The ``specification'' is a bodyless
\texttt{method \{:testEntry\}} signature with no postcondition, plus
a \texttt{Main()} full of \texttt{expect} statements that Dafny
\emph{assumes} (not proves) at verify time. What the verifier
certifies is that these assumed fixtures are consistent enough to
admit a local method-body proof. It is not functional correctness
against a specification. This is faithful to Li et al.'s benchmarked
pipeline, not a replication defect: Figure 6 of \citet{li2025dafny}, reproduced
verbatim in our few-shot prompt, has exactly this shape.

\paragraph{Level 3 is roughly fortyfold smaller than Level 1
($36$--$52\times$ across the three benchmarks), and roughly
benchmark-invariant.} The gate-pass rate stays in a narrow
\textbf{1.9--2.7\% band} across all three benchmarks despite level-2
prevalence doubling from HumanEval (5.3\%) to LCB v6
(10.61\%) as benchmark difficulty rises. Additional
\texttt{ensures} clauses at higher difficulty add inline
postconditions not tied to a named symbol, rather than externalising
specifications. The narrow range suggests the prompt/spec convention
is a stronger limiter than benchmark difficulty; whether it persists
on further benchmarks is a testable prediction. Other structural
(Surface B) and semantic (Surface C) checks are possible (see
\S\ref{sec:gate}, \S\ref{sec:related},
Table~\ref{tab:systems} in Appendix~\ref{app:systems}).

\section{Syntactic binding does not imply semantic fidelity: the $B \to C$ leak in VCG}
\label{sec:vcg}

The $A \to B$ gap in \S\ref{sec:aopus} is stark because Li@Opus's
pipeline puts essentially no structural pressure on the spec. What
about a pipeline that \emph{does} pass our external-spec-binding
instance of Surface B by construction? Does that guarantee Surface C?

We measure this on our own architecture VCG, evaluated on two Rust
benchmarks (MBPP-Rust and VeriContest LC). VCG is a seven-agent
Dafny-as-intermediate pipeline that runs the ensures-binding gate inside the
prover's verify-repair loop and downgrades any component whose
certificate would fail the gate. Across five benchmarks and two
target languages evaluated in this study, VCG ships \textbf{18{,}008
verified components, 100\% of which pass the ensures-binding gate},
i.e.\ 100\% pass our external-spec-binding Surface-B criterion
(other conceivable Surface B instances may still fail on this
population; we do not check those).

Yet some of those shipped-and-structurally-verified samples still
fail hidden tests. This provides operational evidence that Surface C
(semantic specification fidelity) is a distinct failure surface,
under the trusted-toolchain and correct-oracle assumptions stated
below.

\paragraph{Surface C operationalised on the two Rust benchmarks.}
Under the VCG Rust configuration, every component becomes a
verification target and no unverified glue is shipped. Under the
standard trusted-toolchain and correct-oracle assumptions (Dafny
transpilation is sound; the Rust harness runs the intended oracle;
the wrapper/marshalling layer preserves semantics), hidden-test
failure among structurally-verified all-component samples
is a Surface-C proxy. Assigning an individual failure to
specification inadequacy additionally requires excluding implementation,
composition, translation, wrapper, oracle and resource causes. A sample is \emph{fully-verified} iff
every component in its build package shipped as a gate-passing Dafny
certificate.

\paragraph{Two Surface-C instances reported in this section.} Both
condition on samples that have already cleared Surfaces A and B, but
they ask different questions and can catch different failures
(Table~\ref{tab:surfaceC-instances}). Table~\ref{tab:vcg-rust}
reports the first (hidden-test failure) on both Rust benchmarks;
the Post2Exe End2End @1 rate is discussed later in this section on
VeriContest.

\begin{table}[h]
  \caption{Two Surface-C instances reported in this paper. Both
  condition on samples that have already cleared Surfaces A and B.
  They ask different questions and can catch different failure modes.}
  \label{tab:surfaceC-instances}
  \centering
  \small
  \setlength{\tabcolsep}{5pt}
  \begin{tabular}{@{}p{2.2cm}p{4.5cm}p{4.9cm}@{}}
    \toprule
     & \textbf{(i) Hidden-test failure} & \textbf{(ii) Post2Exe} \\
    \midrule
    Question asked
      & Does the \emph{implementation} match the requirement?
      & Does the \emph{postcondition itself} match the requirement? \\
    \addlinespace[2pt]
    Mechanism
      & Run the shipped code against hidden tests
      & Translate the postcondition to an executable predicate; check it accepts positives and rejects negatives \\
    \addlinespace[2pt]
    Reported in
      & Table~\ref{tab:vcg-rust} (both Rust benchmarks)
      & End2End @1 (VeriContest LC) \\
    \bottomrule
  \end{tabular}
\end{table}

The Rust rates on the first operationalisation are in
Table~\ref{tab:vcg-rust}.

\begin{table}[h]
  \caption{VCG on the two Rust benchmarks. \emph{Fully-verified}
  samples pass our Surface-B (ensures-binding) criterion by
  construction and ship a \texttt{dafny verify}-accepted certificate
  on every component. \emph{$B \to C$ failure rate} is the fraction
  of those that nonetheless fail hidden tests, a Surface-C proxy
  under the trusted-toolchain and correct-oracle assumptions.
  \emph{Unconditional failure rate} is the same failure count expressed as
  a share of all attempts (10 samples per problem: 5.96\% on MBPP-Rust,
  3.33\% on VeriContest LC).}
  \label{tab:vcg-rust}
  \centering
  \resizebox{\linewidth}{!}{%
\begin{tabular}{lrrrr}
    \toprule
    Benchmark (Rust) & Total attempts & Fully-verified & $B \to C$ fails & $B \to C$ failure rate \\
    \midrule
    MBPP-Rust \citep{mbpprust}     & 2{,}080 & 1{,}080 & 124 & \textbf{11.48\%} \\
    VeriContest LC \citep{xie2026vericontest} & 6{,}870 & 1{,}545 & 229 & \textbf{14.82\%} \\
    \bottomrule
  \end{tabular}%
}
\end{table}

Expressed as a share of ALL attempts (not conditional), the Rust
residue is small: 5.96\% of MBPP-Rust samples (124/2080) and 3.33\% of
VeriContest-LC samples (229/6870) are fully-verified but fail tests. These are
the two Surface-C proxies reported in this paper; because hidden-test
failure can in principle also reflect transpiler unsoundness,
harness/oracle disagreement, or benchmark-oracle errors, spec
inadequacy is one candidate explanation; these aggregates do not rank
the competing causes.

\paragraph{Why the leak exists even with the gate enforced.} The
ensures-binding gate is an architecture-specific reference check: it
verifies that the certificate \emph{mentions} the spec's predicates.
It cannot verify that those predicates \emph{capture the requirement}.
A spec that names the right predicates but constrains them
incompletely will still ship a structurally-gated certificate whose
impl passes the proof but does not generalise to hidden tests.
Closing that gap requires a semantic check (Post2Exe \citep[\S4.1]{xie2026vericontest} is
one option; surface-grounded spec strengthening, lifting concrete
examples from the problem description into formal lemmas the spec
must satisfy, is another we are pursuing).

\paragraph{Post2Exe as Surface C, applied to VCG on the VeriContest LC
slice.} We ran Post2Exe (the semantic-fidelity check from
\citealt{xie2026vericontest}) on all 1{,}316 VCG-verified+passing
samples on the VeriContest LC slice (687 problems, 10 samples/problem).
\textbf{535 pass all four End2End checks (precond-equivalence,
postcond-Post2Exe, CodeGen, ProofGen), giving an End2End @1 of
7.79\% (535/6870).} The cited external LC rates are 5.51\% (GPT-5.5) and 1.88\%
(single-shot Opus 4.7), giving descriptive contrasts of 2.3 percentage
points and 4.14$\times$. They are external reported comparisons,
not a matched-budget or matched-repair SOTA demonstration. The Post2Exe keep-rate
among verified-and-passing LC samples is 40.7\% (535\,/\,1{,}316).
CF-slice evaluation is left to future work.

\section{Postcondition presence and observed pass rates}
\label{sec:passtax}

Both VCG and Li@Opus show the same pattern in the runs measured here:
samples that carry any \texttt{ensures} clause pass tests
less often than samples without one.

The effect is quantified in Table~\ref{tab:passtax}
(Appendix~\ref{app:passtax}): $-14.0$ pp on HumanEval
($n=86$ vs.\ $1{,}539$), $-6.5$ pp on LeetCodeDataset ($n=213$ vs.\
$1{,}995$), and $-9.4$ pp on LiveCodeBench v6 ($n=1{,}032$ vs.\
$8{,}416$); samples with any \texttt{ensures} clause vs.\ samples
without, respectively.

These are observational associations conditional on generated
postcondition presence. Presence does not establish that a contract is
nontrivial, and the groups need not match in problem difficulty,
generation failures or repair history. The aggregate table therefore
does not identify a causal pass-rate cost of writing specifications.
VCG's reported 3--5 percentage-point difference from BareLLM on Rust
runs is a separate architectural comparison with multiple changing
components, not an identified instance of the same mechanism.

Reporting pass@k alongside structural binding and task outcomes makes
the measured differences visible. A causal explanation requires a
controlled intervention with comparable tasks, model, information,
sampling and repair budgets, and complete failure accounting.

\section{Related work}
\label{sec:related}

\paragraph{LLM-driven verified code generation.} The closest prior
work is \citet{li2025dafny} (Dafny as intermediate language for LLM-driven
verified Python) and VeriContest \citep{xie2026vericontest} (a 946-problem verifiable-Rust
benchmark with a Post2Exe semantic check, our closest Surface-C
analogue). AutoSpec \citep{wen2024enchanting} and Clover \citep{sun2023clover} investigate LLM-generated
Dafny specs specifically. VerusBench \citep{yang2024autoverus}, RepoVBench \citep{zhong2025ragverus},
VeruSAGE \citep{yang2025verusage}, and VeriStruct \citep{sun2025veristruct} evaluate LLM proof generation in
Verus at different scales but do not evaluate end-to-end
NL~$\to$~verified code in the same shape as \citet{li2025dafny} or
VeriContest \citep{xie2026vericontest}.

\paragraph{Certificate integrity and vacuous verification.} Prior
work has noted that verifier acceptance is insufficient:
Dafny's \texttt{\{:axiom\}} and \texttt{\{:verify false\}} attributes
can bypass proof obligations entirely \citep[and community folklore]{dafnydocs},
and any pipeline that ships specs must decide how to handle those
attributes. Our paper generalises: the ``obvious'' vacuity of
\texttt{\{:axiom\}} sits alongside the subtler class of \emph{no
\texttt{ensures} clause at all} (a certificate that proves what
Dafny checks even without any postcondition), which the
ensures-binding gate catches statically. AxDafny \citep{breen2026axdafny} independently
converges on a related family of Surface-B-style checks (missing
\texttt{ensures}, trivially-true postconditions, bypass constructs,
plus an LLM reviewer for other cheating patterns), see
Table~\ref{tab:systems} in Appendix~\ref{app:systems} for a
side-by-side positioning.
Clover \citep{sun2023clover} proposes closed-loop consistency checks among code,
spec, and tests as a related but semantic-rather-than-syntactic
integrity mechanism. VeriContest's Post2Exe check \citep[\S4.1]{xie2026vericontest} is the
closest prior Surface-C instance: compile the emitted postcondition
to an executable predicate and run it against positive/negative
tests. Ensures-binding and Post2Exe are complementary, syntactic
and cheap vs.\ semantic and expensive; a pipeline shipping
certificates ought to report at least one.

\paragraph{Pass@k-only evaluations.} Codex \citep{chen2021evaluating}, AlphaCode \citep{li2022competition}, and
self-consistency / reflection / multi-agent follow-ups \citep{wang2023self,shinn2023reflexion,qian2024communicative}
established the pass@k evaluation methodology on HumanEval \citep{chen2021evaluating},
MBPP \citep{austin2021program}, and LeetCode-style benchmarks. Code-efficiency metrics
\citep{huang2024effibench} also lie on the functional-correctness axis. Our three-surface
decomposition is additive to these, not a replacement.

Our contribution is a metric decomposition, not a new pipeline: the
three-surface decomposition is formalism-agnostic; our ensures-binding
gate is a Dafny-specific Surface-B instance; Post2Exe \citep[\S4.1]{xie2026vericontest} is
a complementary semantic Surface-C instance in Verus. The community
is independently converging on structural (Surface B) checks; this
paper names the surface and reports the paired A-vs-B measurement on
a pipeline that does not enforce a Surface B check itself.
Table~\ref{tab:systems} (Appendix~\ref{app:systems}) places the
closest prior pipelines and evaluations on the three-surface hierarchy.

\section{Conclusion, limitations, and methodological ask}
\label{sec:conclusion}

``Verified'' is not a single property. Our three-surface audit
separates acceptance, external reference and requirement fidelity.
In our same-model reproduction of Li et al.'s pipeline, 99.1\%
Dafny verifier acceptance on HumanEval corresponds to only 1.9\%
external-spec binding, a \textbf{roughly fortyfold} gap holding
across three Python benchmarks of differing difficulty. On our own
architecture that closes this gap by construction, an 11.48\%
(MBPP-Rust) / 14.82\% (VeriContest LC) $B \to C$ gap remains. On the
VeriContest LC slice VCG's End2End @1 is \textbf{7.79\% (535/6870),
a reported result with descriptive external contrasts of +2.3\,pp
against 5.51\% and 4.14$\times$ against 1.88\%; budgets and repair
procedures are not matched}.
\textbf{The methodological ask:} any
pipeline shipping certificates should report at least one of Surface
B (e.g.\ the ensures-binding gate) or Surface C (e.g.\ Post2Exe)
alongside Surface A; where feasible, both, since Surface B alone
leaves the semantic blind spot we demonstrate on VCG, and Surface C
alone leaves the syntactic blind spot we demonstrate on Li@Opus.

\paragraph{Limitations.} \textbf{Single model family:} all primary
results use Claude Opus 4.7. Prompt convention is a plausible explanation of the gap
(\citeauthor{li2025dafny}'s few-shot is a bodyless
\texttt{method \{:testEntry\}} whose \texttt{Main()} runs
\texttt{expect} statements at runtime; the model reproduces that
shape), but the reported comparisons do not isolate it from model
capability or other pipeline differences. Cross-model generalisation (Sonnet 4.7,
GPT-5.5, Gemini 3.1 Pro, open-source models) is open work.
\textbf{One Surface-B instance:} the ensures-binding gate is one
concrete Surface-B check, not the only one; it does not catch
\texttt{predicate P(x: int) \{ true \}} trivialities,
\texttt{\{:axiom\}}-bypassed proofs, or Post2Exe-detectable semantic
gaps \citep[\S4.1]{xie2026vericontest}. AxDafny \citep{breen2026axdafny} converges on a related Surface-B
check that would also work. The methodological ask is ``report
\emph{some} Surface-B or Surface-C instance,'' not specifically this
gate.

\section{AI disclosure section}
In this work, we used generative AI tools for assisting with implementing methods, supporting qualitative and thematic data analysis, creating or editing software code and figures, drafting and editing prose, and checking arithmetic and references. We have reviewed all AI-assisted work; LLM-generated code was verified and tested for correctness by the authors, and we take responsibility for the final content of this work.

\bibliographystyle{iclr2027_conference}
\bibliography{paper-refs} 

\begin{thebibliography}{28}
\providecommand{\natexlab}[1]{#1}
\providecommand{\url}[1]{\texttt{#1}}
\expandafter\ifx\csname urlstyle\endcsname\relax
  \providecommand{\doi}[1]{doi: #1}\else
  \providecommand{\doi}{doi: \begingroup \urlstyle{rm}\Url}\fi

\bibitem[Austin et~al.(2021)Austin, Odena, Nye, Bosma, Michalewski, Dohan,
  Jiang, Cai, Terry, Le, and Sutton]{austin2021program}
Jacob Austin, Augustus Odena, Maxwell Nye, Maarten Bosma, Henryk Michalewski,
  David Dohan, Ellen Jiang, Carrie Cai, Michael Terry, Quoc Le, and Charles
  Sutton.
\newblock Program synthesis with large language models.
\newblock \emph{arXiv preprint arXiv:2108.07732}, 2021.

\bibitem[Breen et~al.(2026)Breen, Letson, Requena~Pozo, and
  Sarra]{breen2026axdafny}
Blair Breen, Alexander Letson, Bego{\~n}a Requena~Pozo, and Luca Sarra.
\newblock {AxDafny}: Agentic verified code generation in {Dafny}.
\newblock \emph{arXiv preprint arXiv:2606.32007}, 2026.
\newblock Validation checks include missing \texttt{ensures}, trivially-true
  postconditions, proof-bypass constructs, and an LLM reviewer for other
  cheating patterns.

\bibitem[Chen et~al.(2021)Chen, Tworek, Jun, Yuan, Pinto, Kaplan,
  et~al.]{chen2021evaluating}
Mark Chen, Jerry Tworek, Heewoo Jun, Qiming Yuan, Henrique Ponde de~Oliveira
  Pinto, Jared Kaplan, et~al.
\newblock Evaluating large language models trained on code.
\newblock \emph{arXiv preprint arXiv:2107.03374}, 2021.

\bibitem[{Dafny Reference Manual}(2024)]{dafnydocs}
{Dafny Reference Manual}.
\newblock {Dafny} documentation: \texttt{\{:axiom\}}, \texttt{\{:verify
  false\}}, \texttt{\{:extern\}} attributes.
\newblock \url{https://dafny.org/latest/DafnyRef/DafnyRef}, 2024.

\bibitem[de~Moura \& Ullrich(2021)de~Moura and Ullrich]{demoura2021lean}
Leonardo de~Moura and Sebastian Ullrich.
\newblock The lean 4 theorem prover and programming language.
\newblock In \emph{Conference on Automated Deduction (CADE-28)}, 2021.

\bibitem[{Emergence AI}(2026)]{mbpprust}
{Emergence AI}.
\newblock {MBPP-Rust}: Rust port of the {MBPP} dataset, 208 problems covering
  task ids 11--510, 2026.
\newblock Released alongside this paper's evaluation artifacts.

\bibitem[Huang et~al.(2024)]{huang2024effibench}
Dong Huang et~al.
\newblock {EffiBench}: Benchmarking the efficiency of automatically generated
  code.
\newblock \emph{arXiv preprint arXiv:2402.02037}, 2024.

\bibitem[Jain et~al.(2024)Jain, Han, Gu, Li, Yan, Zhang, Wang, Solar-Lezama,
  Sen, and Stoica]{jain2024livecodebench}
Naman Jain, King Han, Alex Gu, Wen-Ding Li, Fanjia Yan, Tianjun Zhang, Sida
  Wang, Armando Solar-Lezama, Koushik Sen, and Ion Stoica.
\newblock {LiveCodeBench}: Holistic and contamination free evaluation of large
  language models for code.
\newblock \emph{arXiv preprint arXiv:2403.07974}, 2024.

\bibitem[Lattuada et~al.(2024)Lattuada, Hance, Cho, Brun, Subasinghe, Zhou,
  Howell, Parno, and Hawblitzel]{lattuada2024verus}
Andrea Lattuada, Travis Hance, Chanhee Cho, Matthias Brun, Isitha Subasinghe,
  Yi~Zhou, Jon Howell, Bryan Parno, and Chris Hawblitzel.
\newblock Verus: Verifying {Rust} programs using linear ghost types.
\newblock \emph{Proceedings of the ACM on Programming Languages (OOPSLA)},
  2024.

\bibitem[Leino(2010)]{leino2010dafny}
K.~Rustan~M. Leino.
\newblock Dafny: An automatic program verifier for functional correctness.
\newblock In \emph{Logic for Programming, Artificial Intelligence, and
  Reasoning (LPAR-16)}, 2010.

\bibitem[Li et~al.(2025)Li, Zetzsche, and Somayyajula]{li2025dafny}
Yue Li, Georg Zetzsche, and Siva Somayyajula.
\newblock {Dafny} as verification-aware intermediate language for code
  generation.
\newblock In \emph{Dafny Workshop, co-located with POPL}, 2025.
\newblock arXiv:2501.06283.

\bibitem[Li et~al.(2022)Li, Choi, Chung, Kushman, Schrittwieser, Leblond,
  Eccles, Keeling, Gimeno, et~al.]{li2022competition}
Yujia Li, David Choi, Junyoung Chung, Nate Kushman, Julian Schrittwieser,
  R{\'e}mi Leblond, Tom Eccles, James Keeling, Felix Gimeno, et~al.
\newblock Competition-level code generation with {AlphaCode}.
\newblock \emph{Science}, 2022.

\bibitem[Liu et~al.(2023)Liu, Xia, Wang, and Zhang]{liu2023your}
Jiawei Liu, Chunqiu~Steven Xia, Yuyao Wang, and Lingming Zhang.
\newblock Is your code generated by {ChatGPT} really correct? rigorous
  evaluation of large language models for code generation.
\newblock In \emph{Advances in Neural Information Processing Systems
  (NeurIPS)}, 2023.
\newblock EvalPlus / HumanEval+.

\bibitem[Ma et~al.(2024)]{ma2024specification}
Yifei Ma et~al.
\newblock Specification generation for neural programs via large language
  models.
\newblock \emph{arXiv preprint arXiv:2402.11897}, 2024.

\bibitem[Misu et~al.(2024)]{misu2024towards}
Md. Rakib~Hossain Misu et~al.
\newblock Towards {AI}-assisted synthesis of verified {Dafny} methods.
\newblock \emph{arXiv preprint arXiv:2402.00247}, 2024.

\bibitem[{newfacade}(2024)]{leetcodedataset}
{newfacade}.
\newblock {LeetCodeDataset}.
\newblock \url{https://huggingface.co/datasets/newfacade/LeetCodeDataset},
  2024.

\bibitem[Qian et~al.(2024)Qian, Cong, Yang, Chen, Su, Xu, Liu, and
  Sun]{qian2024communicative}
Chen Qian, Xin Cong, Cheng Yang, Weize Chen, Yusheng Su, Juyuan Xu, Zhiyuan
  Liu, and Maosong Sun.
\newblock Communicative agents for software development.
\newblock In \emph{Association for Computational Linguistics (ACL)}, 2024.

\bibitem[Shinn et~al.(2023)Shinn, Cassano, Berman, Gopinath, Narasimhan, and
  Yao]{shinn2023reflexion}
Noah Shinn, Federico Cassano, Edward Berman, Ashwin Gopinath, Karthik
  Narasimhan, and Shunyu Yao.
\newblock Reflexion: Language agents with verbal reinforcement learning.
\newblock In \emph{Advances in Neural Information Processing Systems
  (NeurIPS)}, 2023.

\bibitem[Sun et~al.(2025)Sun, Sun, Amrollahi, Zhang, Lahiri, Lu, Dill, and
  Barrett]{sun2025veristruct}
Chuyue Sun, Yifan Sun, Dorsa Amrollahi, Eric Zhang, Shuvendu Lahiri, Shan Lu,
  David Dill, and Clark Barrett.
\newblock {VeriStruct}: {AI}-assisted automated verification of data-structure
  modules in {Verus}.
\newblock \emph{arXiv preprint arXiv:2510.25015}, 2025.

\bibitem[Sun et~al.(2023)]{sun2023clover}
Chuyue Sun et~al.
\newblock {Clover}: Closed-loop verifiable code generation.
\newblock \emph{arXiv preprint arXiv:2310.17807}, 2023.

\bibitem[Swamy et~al.(2016)Swamy, Hri{\c{t}}cu, Keller, Rastogi,
  Delignat-Lavaud, Forest, Bhargavan, Fournet, Strub, Kohlweiss, Zinzindohoue,
  and Zanella-B{\'e}guelin]{swamy2016dependent}
Nikhil Swamy, C{\u{a}}t{\u{a}}lin Hri{\c{t}}cu, Chantal Keller, Aseem Rastogi,
  Antoine Delignat-Lavaud, Simon Forest, Karthikeyan Bhargavan, C{\'e}dric
  Fournet, Pierre-Yves Strub, Markulf Kohlweiss, Jean-Karim Zinzindohoue, and
  Santiago Zanella-B{\'e}guelin.
\newblock Dependent types and multi-monadic effects in {F*}.
\newblock In \emph{Principles of Programming Languages (POPL)}, 2016.

\bibitem[Wang et~al.(2023)Wang, Wei, Schuurmans, Le, Chi, Narang, Chowdhery,
  and Zhou]{wang2023self}
Xuezhi Wang, Jason Wei, Dale Schuurmans, Quoc~V. Le, Ed~H. Chi, Sharan Narang,
  Aakanksha Chowdhery, and Denny Zhou.
\newblock Self-consistency improves chain of thought reasoning in language
  models.
\newblock In \emph{International Conference on Learning Representations
  (ICLR)}, 2023.

\bibitem[Wen et~al.(2024)]{wen2024enchanting}
Cheng Wen et~al.
\newblock Enchanting program specification synthesis by large language models
  using static analysis and program verification.
\newblock \emph{arXiv preprint arXiv:2404.00762}, 2024.

\bibitem[Xie et~al.(2026)Xie, Pawagi, Liu, Rai, Shao, Berberian~Jr., Che, and
  Wang]{xie2026vericontest}
Zhiyuan Xie, Mridul Pawagi, Yifei Liu, Ashutosh Rai, Lianmin Shao, Jason
  Berberian~Jr., Sicheng Che, and Wenting Wang.
\newblock {VeriContest}: A competitive-programming benchmark for verifiable
  code generation.
\newblock \emph{arXiv preprint arXiv:2605.08553}, 2026.

\bibitem[Yang et~al.(2024)Yang, Gladshtein, et~al.]{yang2024autoverus}
Chenyuan Yang, Arina Gladshtein, et~al.
\newblock {AutoVerus}: Automated proof generation for rust code.
\newblock \emph{arXiv preprint arXiv:2409.13082}, 2024.
\newblock Introduces VerusBench.

\bibitem[Yang et~al.(2025)Yang, Neamtu, Hawblitzel, Lorch, and
  Lu]{yang2025verusage}
Chenyuan Yang, Norrie Neamtu, Chris Hawblitzel, Jacob~R. Lorch, and Shuvendu
  Lu.
\newblock {VeruSAGE}: A study of agent-based verification for {Rust} systems.
\newblock \emph{arXiv preprint arXiv:2512.18436}, 2025.
\newblock Introduces VeruSAGE-Bench.

\bibitem[Yang et~al.(2023)Yang, Swope, Gu, Chalamala, Song, Yu, Godil, Prenger,
  and Anandkumar]{yang2023leandojo}
Kaiyu Yang, Aidan Swope, Alex Gu, Rahul Chalamala, Peiyang Song, Shixing Yu,
  Saad Godil, Ryan Prenger, and Anima Anandkumar.
\newblock {LeanDojo}: Theorem proving with retrieval-augmented language models.
\newblock In \emph{Advances in Neural Information Processing Systems
  (NeurIPS)}, 2023.

\bibitem[Zhong et~al.(2025)Zhong, Zhu, Tian, and Si]{zhong2025ragverus}
Sicheng Zhong, Jiaqi Zhu, Yifan Tian, and Xujie Si.
\newblock {RAG-Verus}: Repository-level program verification with {LLMs} using
  retrieval augmented generation.
\newblock \emph{arXiv preprint arXiv:2502.05344}, 2025.
\newblock Introduces RepoVBench.

\end{thebibliography}

\appendix

\section{Cascade figure: $A \to B$ vs.\ $B \to C$ collapse}
\label{app:cascade}

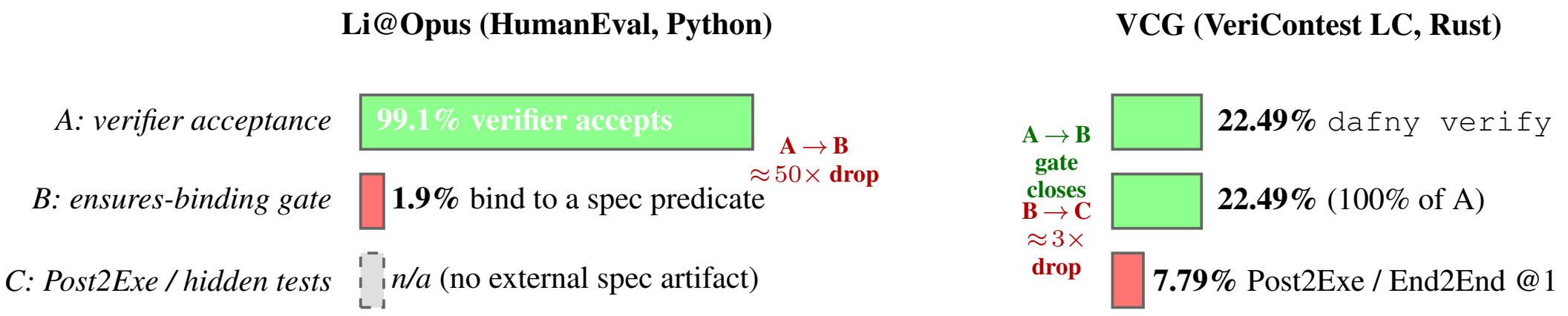
\begin{figure}[t]
  \centering
  \resizebox{\linewidth}{!}{%
\begin{tikzpicture}[
    every node/.style={font=\footnotesize},
    bar/.style={rectangle, draw=black!60, minimum height=0.55cm,
                anchor=west, thick},
    lbl/.style={anchor=west, font=\footnotesize, inner sep=2pt},
    lblIn/.style={anchor=west, font=\footnotesize\bfseries,
                  text=white, inner sep=2pt},
    surflbl/.style={anchor=east, font=\footnotesize\itshape,
                    inner sep=2pt},
    drop/.style={font=\scriptsize\bfseries, text=red!70!black,
                 align=center},
    okmark/.style={font=\scriptsize\bfseries, text=green!45!black,
                   align=center},
  ]

    \node[anchor=south, font=\small\bfseries] at (2.0, 3.30)
      {Li@Opus (HumanEval, Python)};
    \node[anchor=south, font=\small\bfseries] at (9.6, 3.30)
      {VCG (VeriContest LC, Rust)};

    \node[surflbl] at (-0.20, 2.60) {A: verifier acceptance};
    \node[surflbl] at (-0.20, 1.80) {B: ensures-binding gate};
    \node[surflbl] at (-0.20, 1.00) {C: Post2Exe / hidden tests};

    \node[bar, fill=green!45, minimum width=3.964cm] at (0, 2.60) {};
    \node[lblIn] at (0.10, 2.60) {99.1\% verifier accepts};

    \node[bar, fill=red!55,   minimum width=0.15cm] at (0, 1.80) {};
    \node[lbl] at (0.25, 1.80)
      {\textbf{1.9\%} bind to a spec predicate};

    \node[bar, fill=gray!25,  minimum width=0.15cm, dashed] at (0, 1.00) {};
    \node[lbl] at (0.25, 1.00)
      {\emph{n/a} (no external spec artifact)};

    \node[drop] at (4.60, 2.20)
      {A\,$\to$\,B\\$\approx\!50\times$ drop};

    \node[bar, fill=green!45, minimum width=0.900cm] at (7.6, 2.60) {};
    \node[lbl] at (8.60, 2.60)
      {\textbf{22.49\%} \texttt{dafny verify}};

    \node[bar, fill=green!45, minimum width=0.900cm] at (7.6, 1.80) {};
    \node[lbl] at (8.60, 1.80)
      {\textbf{22.49\%} (100\% of A)};

    \node[bar, fill=red!55,   minimum width=0.312cm] at (7.6, 1.00) {};
    \node[lbl] at (7.95, 1.00)
      {\textbf{7.79\%} Post2Exe / End2End @1};

    \node[okmark] at (7.05, 2.20)
      {A\,$\to$\,B\\gate\\closes};
    \node[drop] at (7.05, 1.40)
      {B\,$\to$\,C\\$\approx\!3\times$\\drop};

  \end{tikzpicture}%
}
  \caption{\textbf{Two pipelines fall out of the cascade at different
  surfaces.} Bar widths are proportional to the fraction of
  \emph{attempted samples} surviving each surface (HumanEval: 1640
  attempts; VeriContest LC: 6870 attempts). Li@Opus's collapse is at
  $A \to B$ (99.1\% verifier acceptance, but only 1.9\% carry a
  certificate binding to a spec-defined predicate). VCG closes B on
  100\% of A by construction; its collapse is at $B \to C$
  (22.49\% \texttt{dafny verify} passing to 7.79\% End2End @1). A
  pipeline reporting only Surface A would show 99.1\% and 22.49\%
  respectively for Li@Opus and VCG and hide the shape of failure in both cases.}
  \label{fig:cascade}
\end{figure}

Figure~\ref{fig:cascade} shows the two-pipeline cascade referenced
from \S\ref{sec:intro}.

\section{Li@Opus reproduction details}
\label{app:reprod}

\paragraph{Codebase.} The reproduction consists of prompt files
(verbatim from Figure 6 of \citealt{li2025dafny}), the verify-repair loop
implementation, the compile/shim scripts, and per-benchmark harness
adapters.

\paragraph{Adaptations per benchmark.}

\begin{itemize}
\item \textbf{HumanEval / HumanEval+}: no adaptation; Li et al.'s
prompts assume this shape.
\item \textbf{LeetCodeDataset}: problems reshaped into
function-plus-docstring form (adapter in \texttt{benchmarks.py});
Li's prompts unchanged.
\item \textbf{LiveCodeBench v6}: 611 stdin/stdout problems wrapped
as \texttt{solve(stdin: str) -> str}; 444 function-style problems
required no wrapping. This is an adaptation of Li's pipeline, not
something the paper specifies.
\end{itemize}

\paragraph{Numbers reported (Claude Opus 4.7 via Vertex; Vertex
rejects the temperature parameter for Opus 4.7 so all samples use
default settings).}

\begin{itemize}
\item HumanEval base tests: 92.6\% pass@1 / 97.0\% pass@10
($1{,}519\,/\,1{,}640$ samples).
\item HumanEval+ plus tests: 77.4\% pass@1 / 84.8\% pass@10
($1{,}270\,/\,1{,}640$ samples).
\item LeetCodeDataset: 63.4\% pass@1 / 79.4\% pass@10
($1{,}445\,/\,2{,}280$ samples).
\item LiveCodeBench v6: 55.2\% pass@1 / 76.5\% pass@10
($5{,}819\,/\,10{,}550$ samples).
\end{itemize}

\section{Ensures-binding gate implementation}
\label{app:impl}

The gate is a static AST walk over each generated \texttt{Impl.dfy}
file. Methods with zero \texttt{ensures} clauses fail the gate
immediately (matching the $n \geq 1$ conjunct in \S\ref{sec:gate}).
For every \texttt{method} declaration with at least one
\texttt{ensures} clause, the gate:

\begin{enumerate}
\item Parses \texttt{Impl.dfy} with Dafny's Python-accessible parser
and collects every \texttt{ensures} clause as a Dafny expression AST.
\item Locates the paired \texttt{Spec.dfy} (same component workspace)
and any support files transitively \texttt{include}'d from it.
Enumerates every top-level \texttt{predicate} and \texttt{function}
declaration in these spec files to build the set $\mathrm{Preds}(S)$
of spec-defined predicate/function symbols.
\item For each \texttt{ensures} clause $E_i$, walks its identifier
set and checks whether any identifier resolves to a member of
$\mathrm{Preds}(S)$. If any single identifier does, the clause is
bound; if none of the \texttt{ensures} clauses on the method binds,
the certificate is rejected.
\end{enumerate}

Edge cases handled: transitive \texttt{include} chains; \texttt{function
method}s that are both spec and impl symbols (only the spec-side
declaration counts for $\mathrm{Preds}(S)$); qualified references
(\texttt{Spec.PredicateName}); and \texttt{forall}/\texttt{exists}
bindings whose bound variables don't count against the binding
requirement.

Runtime: under 10 ms per component on our runs.

\section{Existing systems on the three-surface audit}
\label{app:systems}

Referenced from \S\ref{sec:related}.

\begin{table}[h]
  \caption{What existing pipelines and evaluations report on each
  surface. Most systems that report Surface A do not report either
  companion surface.}
  \label{tab:systems}
  \centering
  \small
  \setlength{\tabcolsep}{3pt}
  \resizebox{\linewidth}{!}{%
\begin{tabular}{@{}lcp{4.4cm}p{4.0cm}@{}}
    \toprule
    System & Surface A & Surface B (syntactic binding) & Surface C (semantic fidelity) \\
    \midrule
    \citet{li2025dafny}                    & Yes & Not reported & Runtime fixtures only \\
    VeriContest \citep{xie2026vericontest} end-to-end  & Yes & Not reported & Post2Exe on postcondition \\
    AxDafny \citep{breen2026axdafny}                 & Yes & Missing/trivial \texttt{ensures}, \texttt{\{:axiom\}}, LLM reviewer & Not reported \\
    VCG (our own)                & Yes & Ensures-binding gate & Hidden-test proxy on Rust runs; Post2Exe applied on VeriContest LC (7.79\% End2End @1, 535/6870) \\
    \bottomrule
  \end{tabular}%
}
\end{table}

\section{Pass rates by presence of \texttt{ensures} clause}
\label{app:passtax}

Companion to \S\ref{sec:passtax}.

\begin{table}[h]
  \caption{Pass rate by presence of \texttt{ensures} clause on test-evaluated
  samples, Li@Opus, three Python benchmarks (excluding 9 LeetCodeDataset and
  283 LiveCodeBench v6 compilation/shimming dropouts). Samples with an
  \texttt{ensures} clause pass hidden tests less often than samples without
  one on every benchmark; this observational association among self-selected
  outputs does not identify a causal effect of writing specifications.}
  \label{tab:passtax}
  \centering
  \begin{tabular}{lrrr}
    \toprule
    Benchmark (Python) & With \texttt{ensures} & Without \texttt{ensures} & Delta \\
    \midrule
    HumanEval (base tests)       & 80.2\% ($n=86$)      & 94.2\% ($n=1{,}539$)  & $-14.0$ pp \\
    LeetCodeDataset              & 59.6\% ($n=213$)     & 66.1\% ($n=1{,}995$)  & $-6.5$ pp  \\
    LiveCodeBench v6             & 53.2\% ($n=1{,}032$) & 62.6\% ($n=8{,}416$)  & $-9.4$ pp  \\
    \bottomrule
  \end{tabular}
\end{table}

\end{document}